\documentclass[journal]{IEEEtran}
\usepackage{graphicx}
\usepackage{amsmath}
\usepackage{amssymb}
\usepackage{subfig}
\usepackage{cite}
\begin{document}
\title{Locally Adaptive Hierarchical Cluster Termination With Application To Individual Tree Delineation}
\author{%
Ashlin~Richardson,~\IEEEmembership{Graduate Student Member,~IEEE, }%
\thanks{A. Richardson is with the Department of Mathematics and Statistics, University of Victoria, Victoria, BC, Canada, e-mail: \textbf{ashy@uvic.ca}}%
and~Donald~\thanks{D. Leckie is with the Pacific Forestry Centre, National Resources Canada, Victoria, BC, Canada, e-mail: \textbf{dleckie@nrcan.gc.ca}}%
Leckie\thanks{Prepared April 6, 2011. Submitted to ArXiv.org Nov 30, 2022.}
}
\maketitle
\begin{abstract}
A clustering termination procedure
which is locally adaptive (with respect to the
hierarchical tree of sets representative of the agglomerative merging) is proposed,
for agglomerative hierarchical clustering 
on a set equipped with a distance function.
It represents a multi-scale alternative to conventional scale dependent threshold based termination criteria.

We trim the tree at specific locations by studying cumulative extreme values of rates of change of parameters along paths of the agglomeration hierarchy, each path representing the "history" of successive merges with respect to an initial set. Thus the method considers the smallest localities. 
Moreover, a cumulative extreme value of the rate of a parameter indicates the parameter  
is changing more rapidly than it has yet changed, which, in the context of geometric parameters
we interpret to mean: the geometry at that merging step is 
changing more drastically than at any preceding step.  We refer to this qualitative phenomenon as geometric "paradigm shift".
The method is sensitive to extreme changes in geometry that may happen at any scale.

The proposed termination is presented in the context of a motivating example in forest mapping.
Automated approaches for delineating Individual Tree Crowns (ITCs) from high resolution
 imagery offer the possibility of improving mapping consistency, accuracy, and effectiveness.
The Individual Tree Crown (ITC) methods of Gougeon et al. \cite{gougeon,leckie}
 distinguish individual tree crowns by following the darker "valley" material between them.
Such is a highly effective modus operandi 
for conically shaped trees (softwoods).  Hardwood trees, on the other hand,
 are of greater structural complexity resulting in less distinction between internal and external shadow, 
hence oversegmentation may result.

As a possible avenue for remediating this exceptional situation, the proposed termination method 
offers assistance in detecting instances of oversegmentation by proposing geometrically meaningful candidate groups of segments, from hierarchical clustering using an appropriate "distance".  After developing the approach in this motivating context, we rank the resulting clusters of segments by a scale-independent dispersion measure.\end{abstract}
\section{Motivating Example: Oversegmentation}
\begin{figure}[!t]
\centering
\includegraphics[width=2.5in]{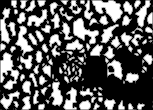}%
\caption{ISOLS (Individual Tree Crown Segments)}
\label{fig1}
\end{figure}
\IEEEPARstart{T}{he} motivating context for the approach is taken to be an Individual Tree Crown (ITC) delineation, 
which specifically refers to a quantity of geospatial regions, or segments,
with each segment ideally corresponding to the crown area of an individual tree. 
Each such segment is known as an isolation (ISOL). The ITC method \cite{gougeon}, based on valley following, 
is most sucessful for coniferous trees, which are largely conical,
so that the tree crown material is 
exhibited as a distinguishable peak, accompanied by the surrounding basal area in the image brightness.  
Hardwood trees are more like broccoli than carrots, showing complicated structure of extrema in the brightness within the tree-top area, making problematic the distinction between basal areas and the darker valley material.
This extreme situation may result in oversegmentation.  In this application, the purpose of the proposed approach is to help identify such a situation by specifying candidate groups of segments whose geometries are consistent with instances of oversegmentation (the "split-case").
\subsection{Algorithmic Overview}
The possible hierarchical relationships between groups of local extrema in the image are, in general, unclear.
How would one distinguish a small group of broccoli plants, from a group of small branches of a large broccoli plant? 
In general it may not be possible to resolve this situation from imagery alone.
The approach considered here is intended to attack the simpler problem: the identification of candidate ISOL groups which potentially represent
an oversegmented tree.
We base the approach upon the following principles: that potential candidate groups of
ISOLS should be spatially compact, though not necessarily circular,
and that some restraints should be placed on their proportions, according
to elementary parameters reflecting their geometry. 

The approach is summarized in three steps: Hierarchical Agglomerative Clustering (HAC) 
using a geometric distance function, a locally adaptive termination procedure for the resulting hierarchy incorporating geometric information,
and finally a ranking of the resulting candidate groups of segments, according to a scale-independent dispersion measure.  
%
\section{Hierarchical Agglomerative Clustering (HAC)}
\begin{figure}[!t]
\centering
\includegraphics[width=2.5in]{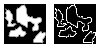}
\caption{At left: (counterclockwise from top right) ISOL areas 1,2,3,4. At right: the ISOL edge pixels.}
\label{fig2}
\end{figure}

For illustrating the distance function used in the Hierarchical Agglomerative Clustering (HAC),
we assume an unreasonably simple input segmentation, consisting of only four ISOLS (Fig. \ref{fig2}). After this illustration, we exhibit case study data featuring a cluster corresponding to the oversegmented case, demonstrating the mechanics of the termination.

The first principle (spatial compactness of candidate groups) we express through Hierarchical Agglomerative Clustering (HAC), 
as applied to groups of ISOLS.  The rich history of HAC includes Wishart's proposal of Hierarchical Mode Analysis \cite{wishart} and  recent generalizations thereof, including those of Stuetzle \cite{Stuetzle2003,stuetzle} and Carlsson \cite{gunnar}.
Standard HAC assumes a choice of distance function between sets. In this example, we choose a geometrically motivated distance, the \textbf{connective distance}, which we define summararily.  This distance is intended to represent the darker valley area corresponding to the interstitial 
matter between the segments.
\subsection{Connective Links Using Compass Directions}
For each ISOL in the image, we locate the boundary (edge) pixels. For a given ISOL, we find \textbf{connective
links} with adjacent ISOLS according to the following heuristic: from a given edge pixel,
we consider a path emanating in each of the eight compass directions.
The path continues in a straight line across interstitial pixels not associated with any ISOL.
Supposing the path reaches a pixel corresponding to an ISOL other than
the originating ISOL, the path is deemed a \textbf{connective
link} between the two ISOLS. We denote the set of \textbf{connective links}
between A and B as $L(A,B)$ so that $l\in L(A,B)$ represents a given
 link, and $p\in l$ represents a pixel representative of the connective
link. Also, for a given ISOL $A$ we will have occasion to use the pixels
representative of $A$, which we will denote by $P(A).$ The
edge pixels belonging to $A$ we will refer to by $E(A).$ 
\subsection{Connective Distance Between ISOLS}
\begin{figure}[!t]
\centering
\includegraphics[width=2.5in]{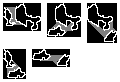}
\caption{Connective area between ISOL pairs:\\(1-2), (1-3), (1-4), (2-3), (3-4) respectively.}
\label{fig3}
\end{figure}

We compute the connective links between any applicable pairs of ISOLS,
making possible the definition of 
the \textbf{connective distance} between any two ISOLS, 
using $L(A,B)$. 
Since an individual connective link $l \in L(A,B)$ is comprised of the representative pixels,
we define the pixel count $|L(A,B)|$ corresponding to this collection of links 
as the count of the pixels within the union $|\cup L(A,B)|$ of the individual links from A to B (or from B to A).
The \textbf{connective distance} $d(A,B)$ is taken to be this pixel count, formally:
\[
d(A,B)=|L(A,B)|=|\cup L(A,B)|.
\]
In Fig. \ref{fig3} we show the \textbf{connective area} for the example, in grey (the ISOL edges are in white).  Next it is necessary to define the connective area between two groups
of ISOLS. 
\subsection{Connective Distance Between Groups Of ISOLS}
Given two groups of isols $\mathbb{A}$ and $\mathbb{B}$, we define
the \textbf{connective area} between them as the number of pixels
in the union of the set $L(A,B)$ of \textbf{connective links} between any pair of ISOLS $A,B$
with $A$ in group $\mathbb{A}$ and $B$ in the other group $\mathbb{B}$:
\begin{align}
d(\mathbb{A},\mathbb{B})=\big|\underset{A\in\mathbb{A},B\in\mathbb{B}}{\cup}L(A,B)\big|.
\label{connective_distance}
\end{align}
We should also define the set of \textbf{connective links} between two groups,
say group $\mathbb{A}$ and group $\mathbb{B}$, as the set of all
connective links between any pair of ISOLS $A$ and $B$, with ISOL $A$ in group
$\mathbb{A}$ and ISOL $B$ in group $\mathbb{B}$:
\[
L(\mathbb{A},\mathbb{B})=\underset{A\in\mathbb{A},B\in\mathbb{B}}{\cup}L(A,B).
\]
\subsection{Hierarchical Agglomerative Clustering}
Each iteration of the Hierarchical Agglomerative Clustering (HAC)
operates upon a list $(S)$ of groups of ISOLS. The list is initialized
with groups consisting of the individual ISOLS:
\[
S=\{\{ISOL_{1}\},\{ISOL_{2}\},\dots\}.
\]
Each iteration of the HAC simply merges together the two closest groups
in the list $S$ (closest in the sense of (\ref{connective_distance}), the \textbf{connective distance}
between groups of isols). 

1) The two closest groups are: 
\[
(\mathbb{A},\mathbb{B})=\underset{\mathbb{A}\in S,\mathbb{B}\in S,A\neq B}{argmin}\{d(\mathbb{A},\mathbb{B})\}.
\]

\begin{figure}[!t]
\centering
\includegraphics[width=2.5in]{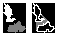}
\caption{Left: the sets joined at the first iteration. Right: the associated connective area between groups (grey).}
\label{fig4}
\end{figure}

\begin{figure}[!t]
\centering
\includegraphics[width=2.5in]{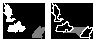}
\caption{Left: the sets joined at the second iteration. Right: the associated connective area between groups (grey).}
\label{fig5}
\end{figure}

\begin{figure}[!t]
\centering
\includegraphics[width=2.5in]{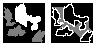}
\caption{Left: the sets joined at the third iteration.\\ Right: the associated connective area between groups (grey).}\label{fig6}
\end{figure}

2) Upon determining the closest $\mathbb{A},\mathbb{B}$ in the list
$S,$ the groups are united (where the $=$ sign means assignment):
\[
\mathbb{A}=\mathbb{B}=\mathbb{A}\cup\mathbb{B}.
\]
Then we return to 1), repeating until all groups are united.
In this application, the end result is the group of all ISOLS.
Supposing there are $N$ iterations indexed by $i=1,\dots,N$
we denote, while developing in the next section the \textbf{agglomeration hierarchy}, 
the merged result $\mathbb{A}\cup\mathbb{B}$ at
iteration $i$ as: 
\[
h_{i}=\mathbb{A}\cup\mathbb{B}.
\]
The merging for the trivial example is shown in Fig. (\ref{fig4}-\ref{fig6}).
%
\subsection{The Agglomeration Hierarchy}
\begin{figure}[!t]
\centering
\includegraphics[width=2.5in]{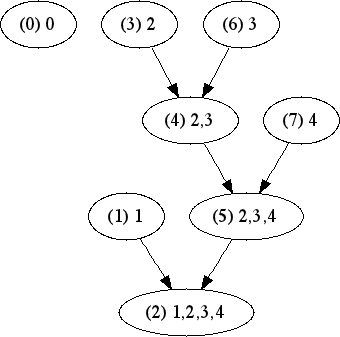}
\caption{An example \textbf{agglomeration hierarchy}.  Arrows represent the \textbf{edge set}; ellipsoids represent the \textbf{vertex set}.}
\label{fig7}
\end{figure}

The \textbf{agglomeration hierarchy} is the list of all groups that have ever
been part of the list $S$, that is, any groups of ISOLS that existed in any step of the
merging procedure; the associations between such groups indicate the
ancestral history of each, detailing exactly which groups of ISOLS are merged. 
Considering the ISOL groups in the history to correspond directly to the nodes of the graph (the \textbf{vertex set})
we view the agglomeration hierarchy as a graph.
Ancestral relationships between the groups merged are taken to correspond with the links between nodes on the graph - ordered pairs (parent, child) of ISOL groups.
In graph terminology, these links comprise the \textbf{edge set}.  For the trivial example, Fig. (\ref{fig7}) shows the simple hierarchy which results from aggolomeration.

Because the merging started with the list of singleton groups corresponding
to individual ISOLS, each such group $\{ISOL_{j}\}$ where $j=1,\dots,M$  ($M$ is the number
of ISOLS) is a member of the vertex set of the hierarchy.
As any group in the hierarchy corresponds to a vertex (node) of
the graph, each $h_{i}$, i.e. the merging result at each
iteration $i$, also corresponds to a vertex (node). The last merging
result is the group consisting of all of the ISOLS in the image, i.e., 
\[
h_{N}=\{ISOL_{1},ISOL_{2},\dots,ISOL_{M}\}.
\]
Again, the links between nodes of the graph may be represented as ordered pairs.
Supposing at some stage the groups $\mathbb{A}$ and $\mathbb{B}$
are merged to form $h_{i}=\mathbb{A}\cup\mathbb{B}$, then we add
to the edge set of the hierarchy the two edges $(\mathbb{A},h_{i})$ and $(\mathbb{B},h_{i})$.
That is, graphically, lines or arrows drawn from both $\mathbb{A}$
and $\mathbb{B}$ to $h_{i}$ indicate that $\mathbb{A}$ and $\mathbb{B}$
were merged to form $h_{i}$. 
\subsection{Linear Paths On The Agglomeration Hierarchy}
\subsubsection{Immediate Successors }
Examining the history of an individual ISOL, $\{ISOL_{j}\}$,
in terms of its membership in successively larger groups that
form, we start at $\mathbb{A}=\{ISOL_{j}\}$
simply following the edges of the graph to the next merged group involving
$\mathbb{A},$ (the edge set has a unique
edge $(\mathbb{A},h_{i})$ starting at $\mathbb{A}$ and ending at
$h_{i}$). Because any node $\mathbb{A}$ in the hierarchy has one
successor, that is, the group $\mathbb{A}$ is merged with the nearest
other group ($\mathbb{B}$) in the list $S$ to form $h_{i},$ in
general for a group of ISOLS $\mathbb{A}$ it seems handy to reserve
the notation $h(\mathbb{A})$ to represent the "successor" of
$\mathbb{A}$. So a linear path from $\{ISOL_{j}\}$ can be written as:
\begin{align}
\{\{ISOL_{j}\},h(\{ISOL_{j}\}),h(h(\{ISOL_{j}\})),\dots, \nonumber \\
h(h(\dots h(\{ISOL_{j}\})\dots))=h_{N}\}.  \nonumber
\end{align}
That is, starting with the ISOL, we hop down node by node (by repeated applications of $h()$) until
we reach the bottom:
\[
h(h(\dots h(\{ISOL_{j}\})\dots))=h_{N}.
\]
Thus, the terminal node $h_{N}$ has no successors:
\[
h(h_{N})=\{\}.
\]
\subsubsection{Immediate Ancestors}
In later sections we find it useful to denote the set of \textbf{immediate
ancestors} of a group by $h^{-1}.$ That is, $h^{-1}(\mathbb{A})$
represents the two (or zero) groups merged to form the group $\mathbb{A}$.
So, $h^{-1}(\mathbb{A})$ either has two elements since it resulted from a merge, or $h^{-1}(\mathbb{A})=\{\}=\emptyset$,
i.e., having no ancestors, the group consists of a single
ISOL. Supposing further that $H$ is a list of different nodes in
the hierarchy, we extend the definition of ancestor set to apply to
$H$, so that we can find the \textbf{immediate ancestors} of all
 elements of $H$, rather than just those of a single element:
\[
h^{-1}(H)=\{h^{-1}(h_{\lambda})\,|\, h_{\lambda}\in H\}.\]
The analogous definition for the \textbf{immediate successors} of
a list of groups is:
\[
h(H)=\{h(h_{\lambda})\,|\, h_{\lambda}\in H\}.
\]
\subsubsection{All Successors And Ancestors For A Given Hierarchical Node}
The above definitions allow us to refer to \textbf{successors at depth}
$k$ as: 
\[
h^{k}(H)=\underset{h_{i}\in H}{\cup}\{h(\dots h(h_{i}))\}\]
($k$ applications of $h$) and the \textbf{ancestors at height} $k$
by: 
\[h^{-k}(H)=\underset{h_{i}\in H}{\cup}\{h^{-1}(\dots h^{-1}(h_{i}))\}\]
($k$ applications of $h^{-1}$) to define the \textbf{set
of all successors}:\[
h^{\infty}(H)=\{H\}\cup\overset{\infty}{\underset{k=1}{\cup}}\{h(H)\}\]
and the \textbf{set of all ancestors}:
\[
h^{-\infty}(H)=\{H\}\cup\overset{\infty}{\underset{k=1}{\cup}}\{h^{-k}(H)\}.\]
Supposing $H$ is a node, the set of all successors
of $H$ is the linear path beginning at $H$ and ending at the terminal
node representing all the ISOLS in the image. For a given 
element $h_i$ of the hierarchy, shortly we will have occasion to refer to the pixels representative of that group: 
\begin{align}
	P(h_i) = \underset{A \in \cup h^{\infty}(h_i)}{\cup} P(A). \label{PHI}
\end{align}
This is possible from combining the pixels $P(A)$ from ISOLS belonging to any of the ancestor groups.
\section{Termination Via Geometric Parameter Rates}
\subsection{Possible Geometric Parameters Of Interest}
\subsubsection{$A_{merge}$ }
By $A_{merge}$ we refer to the connective area $d(\mathbb{A},\mathbb{B})$
when ISOL groups $\mathbb{A}$ and $\mathbb{B}$ are merged to form
$h_{i}=\mathbb{A}\cup\mathbb{B}$. 
We can also think of this as $P(h_i)$ as above (\ref{PHI}).
\subsubsection{$LW_{ratio}$ }
The parameter $LW_{ratio}$ represents the ratio
between "length" and "width" of a hypothetical pixel region corresponding to
the connective area $d(\mathbb{A},\mathbb{B})$ when ISOL groups $\mathbb{A}$
and $\mathbb{B}$ are merged to form $h_{i}=\mathbb{A}\cup\mathbb{B}$,
assuming that such a hypothetical pixel region corresponding to the connective area
is in fact rectangular - of course, this is not the case.
Yet the parameter necessarily captures the interplay between "length" and "thickness" of the connective area in an analogous sense. In particular, we calculate the estimated "length" of the connective
area associated with the merge as:
\[
\hat{l}=\frac{1}{|L(\mathbb{A},\mathbb{B})|}\underset{\mathbb{L}\in L(\mathbb{A},\mathbb{B})}{\sum}|\mathbb{L}|.
\]
That is, taking the total length of connective links between groups:
$\mathbb{A}$ and $\mathbb{B}$, that is, $\underset{\mathbb{L}\in L(\mathbb{A},\mathbb{B})}{\sum}|\mathbb{L}|$,
dividing this by the total number of links between $\mathbb{A}$
and $\mathbb{B}$, that is, $|L(\mathbb{A},\mathbb{B})|$, resulting
in $\hat{l}$, we arrive at the average length of a link between the ISOL groups
$\mathbb{A}$ and $\mathbb{B}$. Assuming such a region is rectangular, we estimate the "width"
of the hypothetical region as:
\[
\hat{w}=\frac{A_{merge}}{\hat{l}}.
\]
Then the hypothetical ratio between "length" and "width" for the region is:
\[
LW_{ratio}=\frac{\hat{l}}{\hat{w}}=\hat{l}/\frac{A_{merge}}{\hat{l}}=\frac{(\hat{l})^{2}}{A_{merge}}.
\]
\subsubsection{$N_{pix}$ }
The parameter $N_{pix}$ represents the total area of ISOLS
belonging to the group formed by the current merge.
Supposing the current hierarchical node (a group of ISOLS) is $H,$
$\cup H$ refers to the ISOLS within the group $H.$ Then the total ISOL
area corresponding to $H$ is the number of pixels from component ISOLS:
\[
\underset{A\in H}{\sum}P(A).
\]
\subsubsection{$N_{edge}$ }
The parameter $N_{edge}$ represents the perimeter, as the count of all edge pixels,
 of ISOLS belonging to the
group formed by the current merge. Then the total
perimeter of ISOLS corresponding to $H$ is the  number of edge
pixels corresponding to all component ISOLS:
\[
\underset{A\in H}{\sum}E(A).
\]
\subsubsection{$A_{cumulative}$ }
The parameter $A_{cumulative}$ represents the interstitial area corresponding
to the group of ISOLS formed by the current merge: $A_{cumulative}$ is the pixel count corresponding to the combined connective areas
pertaining to merges ancestral to (and including) the current merge. We define the \textbf{connective links}
of the hierarchical element $h_{i}$ by: 
\[
L(H)=\underset{\mathbb{A}\in h^{-1}(H),\mathbb{B}\in h^{-1}(H),\mathbb{A}\neq\mathbb{B}}{\cup}L(\mathbb{A},\mathbb{B}).
\]
This allows us to define the cumulative set of connective links,
by following the hierarchy "upwards" by listing all ancestors $h^{-\infty}(H)$
of the current node $H$, 
combining together the connective links associated with all such ancestors:
\[
\underset{h_{i}\in h^{-\infty}(H)}{\cup}L(h_{i}).
\]
Combining and counting the pixels from all such links, we
arrive at the combined area from the connective links 
of all merges in the history of the node $H:$ 
\[
A_{cumulative}  = \bigg|\cup\bigg[\underset{h_{i}\in h^{-\infty}(H)}{\cup}L(h_{i})\bigg]\bigg|.
\]
\subsection{Geometric Parameters On Linear Paths}
Along the linear path originating from $\{ISOL_{j}\}$: 
\begin{align}
\{\{ISOL_{j}\},h(\{ISOL_{j}\}),h(h(\{ISOL_{j}\})),\dots,  \nonumber \\
h(h(\dots h(\{ISOL_{j}\})\dots))=h_{N}\}   \nonumber
\end{align}
we can plot any of the above parameters, or ratios thereof. This is
a locally adaptive method; in particular, the locality considered is the starting ISOL itself.
 The locality expands (with
successive merges) along the linear path, until the terminal node
$h_{N}$ is reached. Therefore the linear path allows 
a multi-scale study of the clustering behaviour, from local to completely global. 
We watch for transitions between progressively greater scales by performing
calculus of the parameters upon linear paths on the hierarchy. The
length of a path starting at node $H$ is the size of the set of
successors:
\[
\big|h^{\infty}(H)\big|.
\]
\subsection{Rates Of Geometric Parameters Along Such Linear Paths}
Given one of the parameters above (or ratios thereof) 
we examine the values along the above linear path. 
In particular, finite differences of such parameters along the linear
path provide valuable information. We suppose a parameter takes a sequence
of values along the linear path as follows:
\[
f_{0},f_{1},f_{2},\dots,f_{\big|h^{\infty}(H)\big|}=\{f_{j}\}_{j\in0,\dots\big|h^{\infty}(H)\big|},
\]
where the nodes along the path from the node $h_{0}$ are:
\[
h_{0},h(h_{0}),h(h(h_{0})),\dots h^{\big|h^{\infty}(H)\big|}(h_{0})=\{h_{j}\}_{j\in0,\dots\big|h^{\infty}(H)\big|}.
\]
For a given node $H$ we denote by $i(H)$ the
merge iteration at which $H$ was formed (not defining  $i(H)$ for $H$ without ancestors), and if in addition, the starting element $h_{0}$ is
understood, we denote by $j(H)$ the j-index of $H=h_{j}$ in
the linear sequence starting at $h_{0}.$ We can define $j^{th}$
differences in two ways, as follows:
\begin{align}
D_{j}&=\frac{f_{j+1}-f_{j}}{i(h_{j+1})-i(h_{j})}  \label{D1} \\
D_{j}&=\frac{f_{j+1}-f_{j}}{j+1-j}=f_{j+1}-f_{j}.  \label{D2} 
\end{align}
The first is interesting, as it expresses the rate of change
of the parameter $f$ with respect to the iteration number $i$; this could be the subject of further research. In this implementation
we use (\ref{D2}), expressing the rate of change of
the parameter $f$ with respect to the index of the location within
the linear path. 
\subsection{Extrema Of Rates Of Geometric Parameters}
We apply the following view: when
considering a location on a path being followed,
an extreme value of greater magnitude than encountered thus far
 (according to the rate of change of the parameter $f$)
should indicate that a qualitative change 
has occurred at the merge corresponding to the path location considered.
We use the first order difference (\ref{D2}) noting the use of higher
order differences as a possible subject of further inquiry. 
The current implementation scales the parameter in the range $[0,1]$, calculating (\ref{D2}) 
for $j=0$ to $\big|h^{\infty}(H)\big|-1$. Then we take the \textbf{cumulative maximum} of (\ref{D2}) as:
\begin{align}
C_{max}(J)=\max\{D_{j}\,|\, j\leq J\}.   \label{cumsum}
\end{align}
A point $j$ where $C_{max}(j)\neq C_{max}(j-1)$ is deemed a \textbf{break point}, representing a candidate location indicative of a substantiative qualitative change in the merging behaviour. We denote the set of break points for (\ref{cumsum}) by:
\[
B_{h_{0}}=\{j\,|\, C_{max}(j)\neq C_{max}(j-1)\},
\]
the indices along the linear sequence where the "upward
steps" in the transformed sequence (\ref{cumsum}) are observed. 
\subsection{Histogram Thresholding Of Break Point Counts}
Supposing $H$ represents all nodes within the hierarchy, and
$H_{0}$ represents the singleton nodes within the hierarchy:
given a starting element $h_{o}$, we denote by $\chi_{H}(B_{h_{0}})$
the indicator function for the breakpoints of the transformed sequence (\ref{cumsum}) 
over the path starting at $h_{0}.$ That is, $\chi_{H}(B_{h_{0}}) \big|_{h}=1$
 for a given hierarchical node $h$
if $h\in B_{h_{0}}$, and is $0$ otherwise. That is, for a given node, $\chi_{H}(B_{h_{0}})$ indicates whether or not the
transformed sequence has a break point at the given node.
Then, we count (at each node) all break points
recommended by all the individual paths beginning at singleton
nodes:
\[
\underset{h_0\in H_{0}}{\sum}\chi_{H}(B_{h_{0}}) .
\]
Evaluating this function at a node $h$ gives a count of break
points experienced at $h$, from all paths considered:
\[
F(h) = \underset{h_0\in H_{0}}{\sum}\chi_{H}(B_{h_{0}}) \big|_{h} .
\]
We evaluate the histogram of values $\{F(h)\,|\, h\in H\}$ (shown in Fig. 12 for the case study) obtained
by $F(h)$ to perform a significance analysis, setting an arbitrary threshold $p=25\%$ to obtain the results shown. 
The significance value is taken to be $F_{significance}$ for which $p=25\%$ of the
values of the histogram of $F(h)$ are to the right of $F_{significance}$.

We consider hierarchical nodes with a break point count above
$F_{significance}$ to be 
indicative of the desired "paradigm shift", or change in qualitative
merging behaviour (with respect to the parameters of interest).  
Accordingly, we trim the hierarchy by nullifying any
nodes (and their successors) whose break point count is above $F_{significance}$.
The terminal nodes  $h^{\infty}(H_{0})$   of the trimmed hierarchy  $H'$ are the computational output of this approach. 
Since the termination trims
the hierarchy giving attention to the location of trimming
according to the collective behaviour of the individual linear paths,
the approach is locally adaptive to the hierarchical structure resulting from hierarchical agglomeration.
\section{Case Study}
\begin{figure}[!t]
\centering
\includegraphics[width=2.5in]{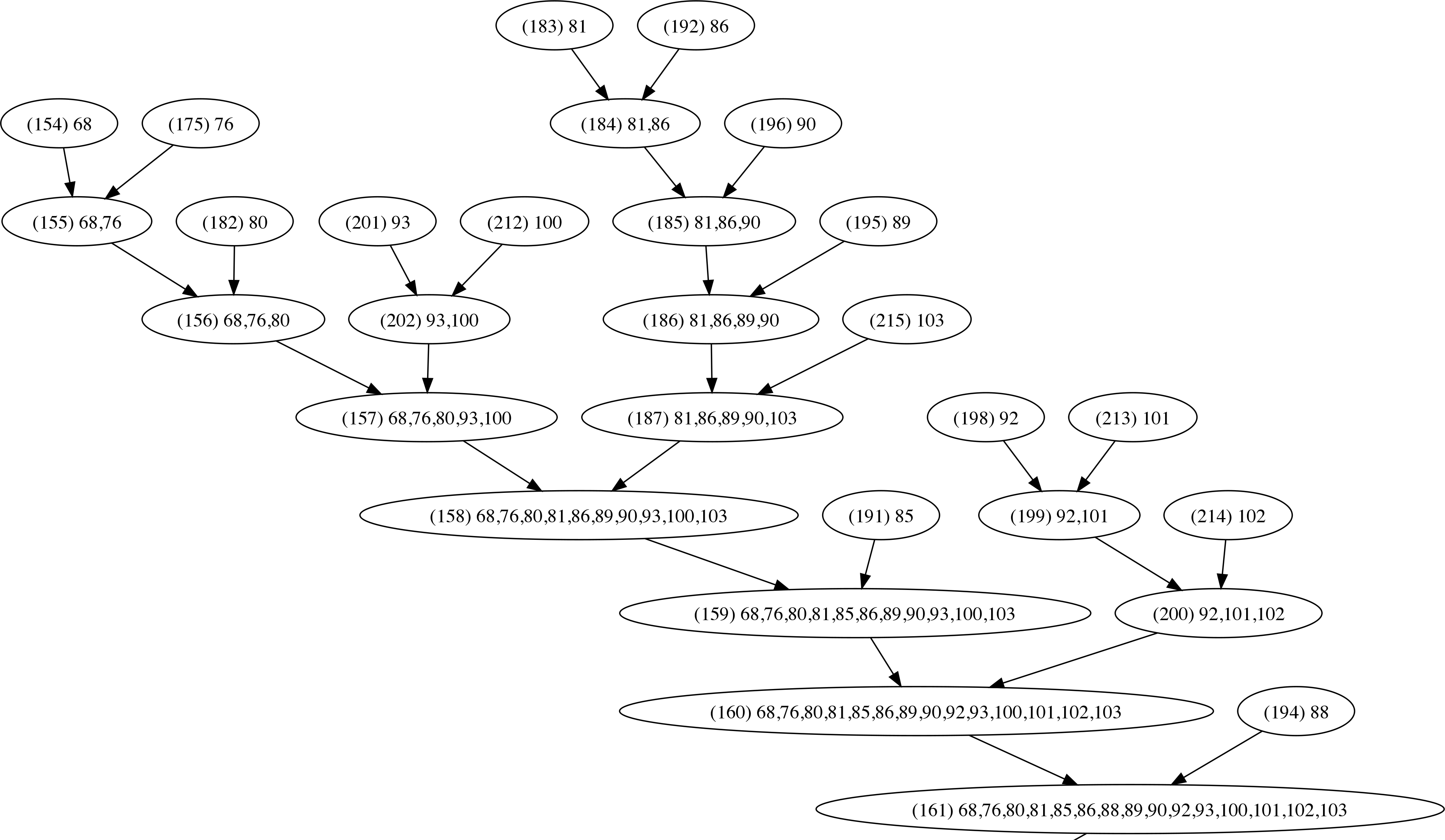}%
\caption{A small component of the hierarchy for the case study.}
\label{fig8}
\end{figure}

\begin{figure*}[!t]
\centerline{
\subfloat[$A_{merge}$]{
\includegraphics[width=2.5in]{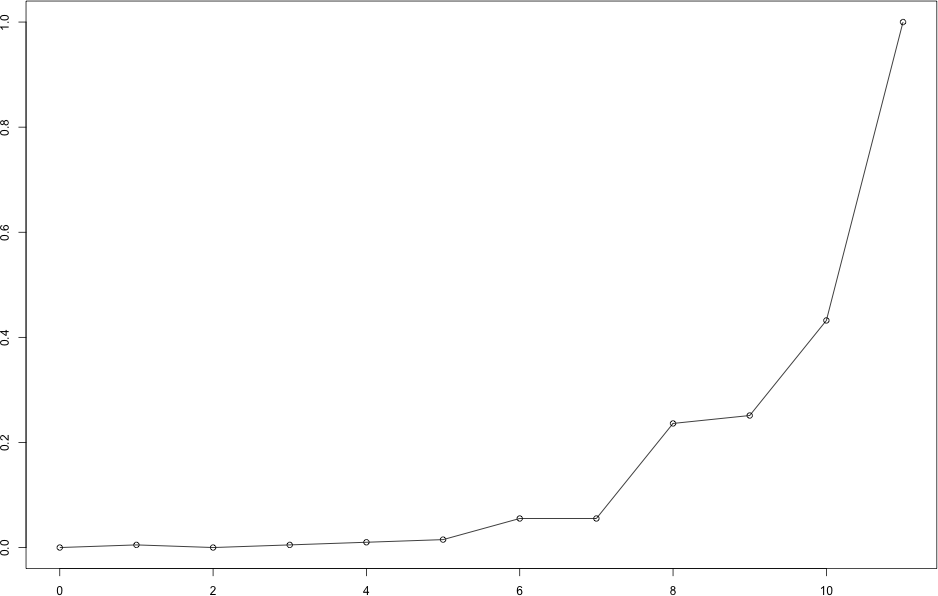}%
\label{asdf}}
\hfil
\subfloat[ $\frac{LW_{ratio}}{A_{cumulative}}$ ]{
\includegraphics[width=2.5in]{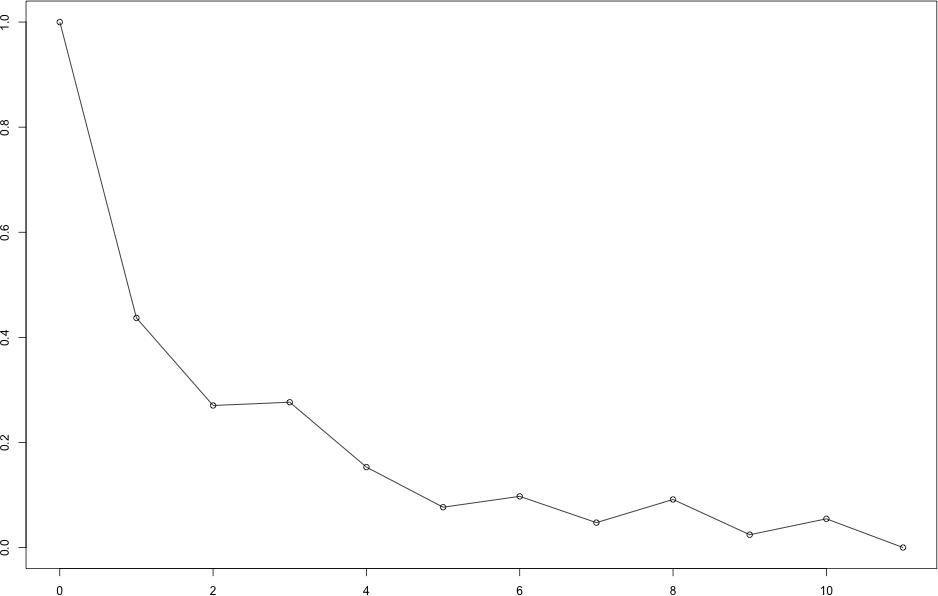}%
\label{fig_second_case}}}
\caption{Along the linear path starting at ISOL 103: the parameter without transformations.}
\label{fig9}
\end{figure*}

The case study is the area of 131 ISOLS shown in Fig. \ref{fig1}.
The agglomeration hierarchy for this example is complicated, so a small sample is shown in Fig. \ref{fig8}.
For the termination procedure, we plot the parameters and results corresponding to both $A_{merge}$ and $\frac{LW_{ratio}}{A_{cumulative}}$.  Termination using either parameter successfully isolated the desired feature: the circular ISOL arrangment near the image centre.
In Fig. (\ref{fig9}) the parameters $A_{merge}$ and $\frac{LW_{ratio}}{A_{cumulative}}$  are shown along
a path beginning at ISOL 103, which is inside the desired image feature.
Next, transformations of $A_{merge}$ and $\frac{LW_{ratio}}{A_{cumulative}}$ are shown: the first difference (\ref{D2}) in Fig. (10) and the cumulative maximum of the first difference (\ref{cumsum}) in Fig. (11).  Again, break points are determined according to the "steps" exhibited by the graph of (\ref{cumsum}). This procedure calculates the transformed parameter along all trajectories starting at the initial hierarchical nodes $H_{0}$ - in each case, the local indices along the linear path are converted into global (nonlinear) indices representing the hierarchical elements themselves, in order that a count of break points may be accumulated over the entire hierarchy $H$.

Once the terminal nodes of the hierarchy are obtained from the adaptive termination procedure, 
we discard those representing groups of ISOLS numbering below a threshold: for this study, we chose 7.  Such a choice limits the display of groups that do not represent substantial clusters. 
\begin{figure*}[!t]
\centerline{
\subfloat[$A_{merge}$]{
\includegraphics[width=2.5in]{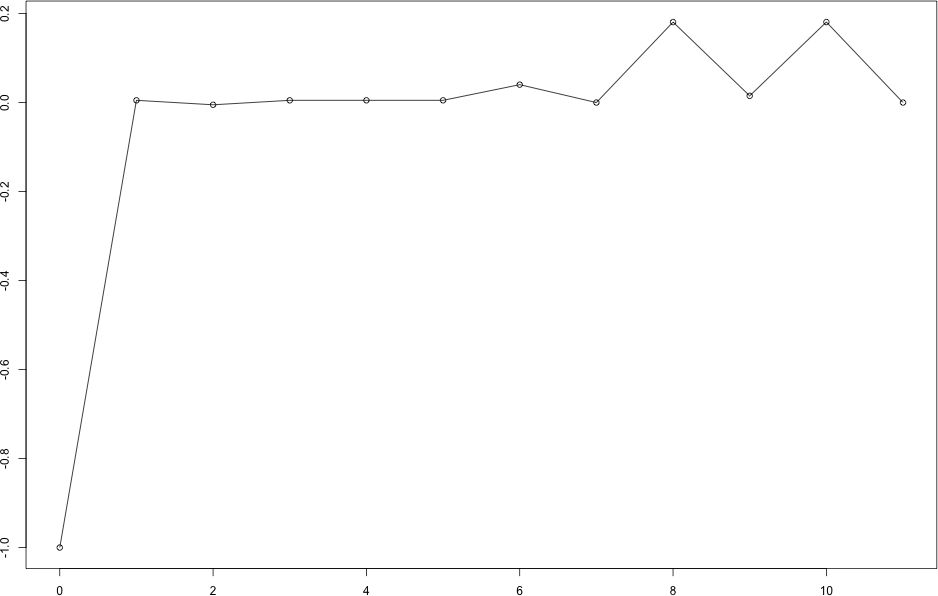}%
\label{fig10}}
\hfil
\subfloat[$\frac{LW_{ratio}}{A_{cumulative}}$ ]{
\includegraphics[width=2.5in]{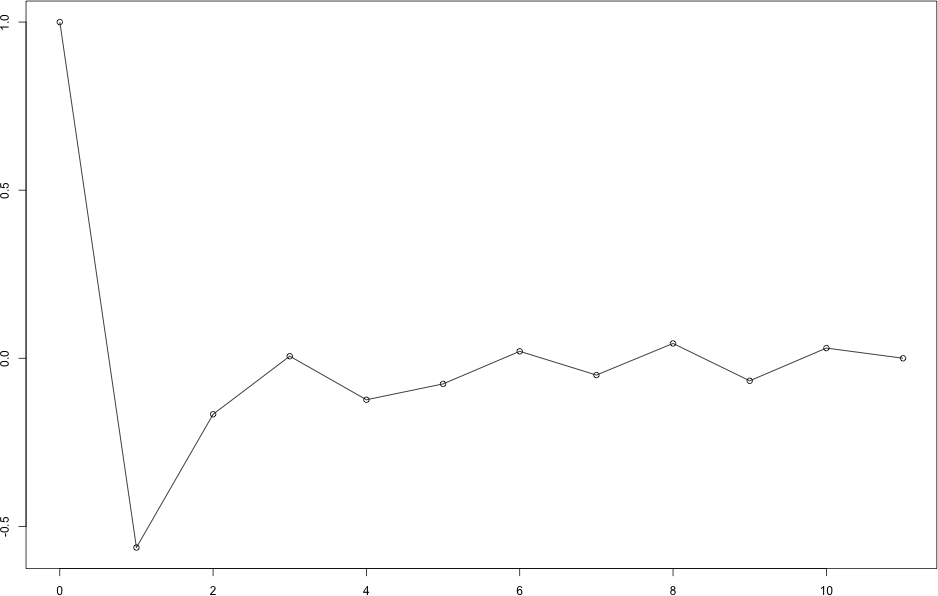}%
\label{fig_second_case}}}
\caption{Along the linear path starting at ISOL 103: the first difference of the parameter.}
\label{fig11}
\end{figure*}
\begin{figure*}[!t]
\centerline{
\subfloat[$A_{merge}$]{
\includegraphics[width=2.5in]{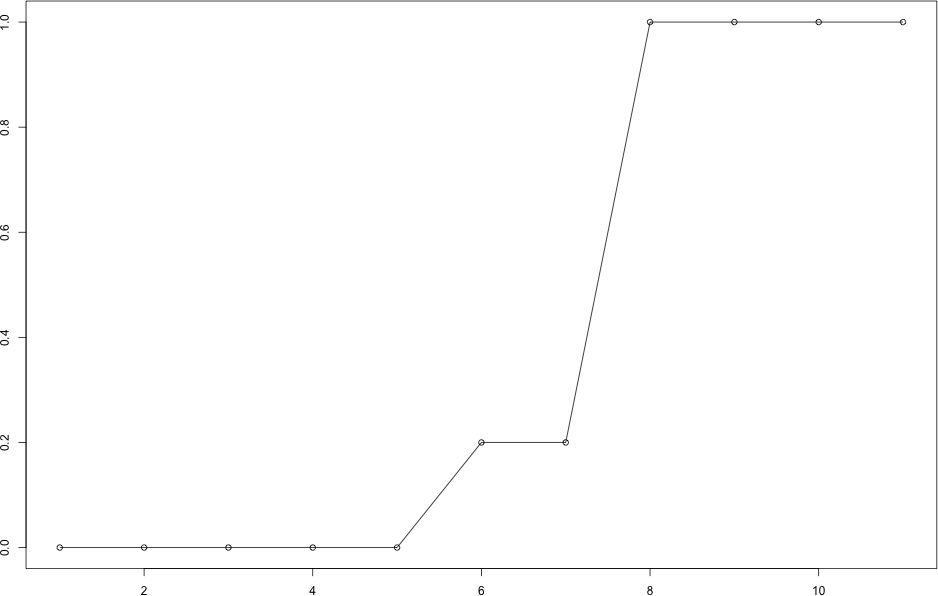}%
\label{fig12}}
\hfil
\subfloat[$\frac{LW_{ratio}}{A_{cumulative}}$ ]{
\includegraphics[width=2.5in]{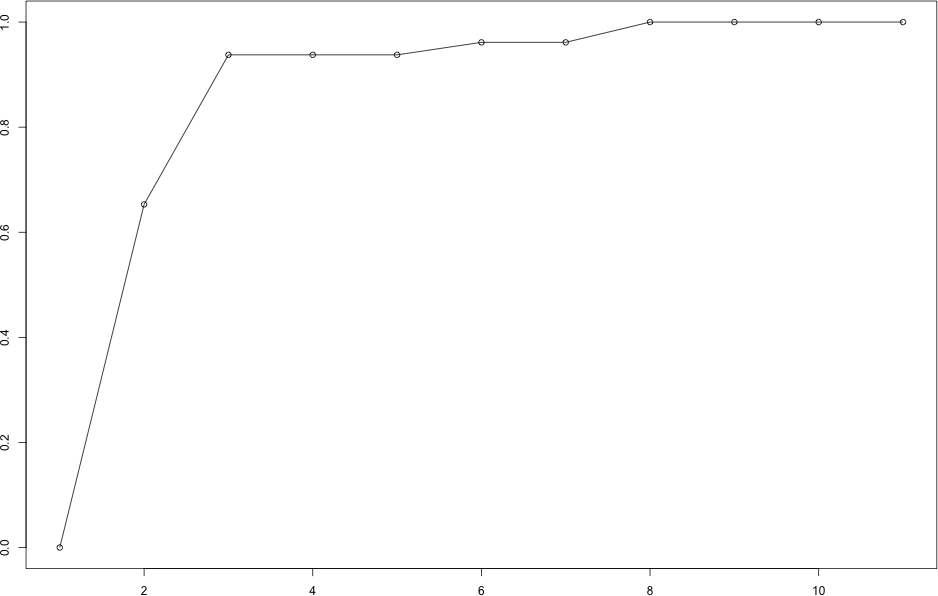}%
\label{fig_second_case}}}
\caption{Along the linear path starting at ISOL 103: the cumulative maximum of the first difference.}
\label{fig13}
\end{figure*}
\begin{figure*}[!t]
\centerline{
\subfloat[$A_{merge}$]{
\includegraphics[width=2.5in]{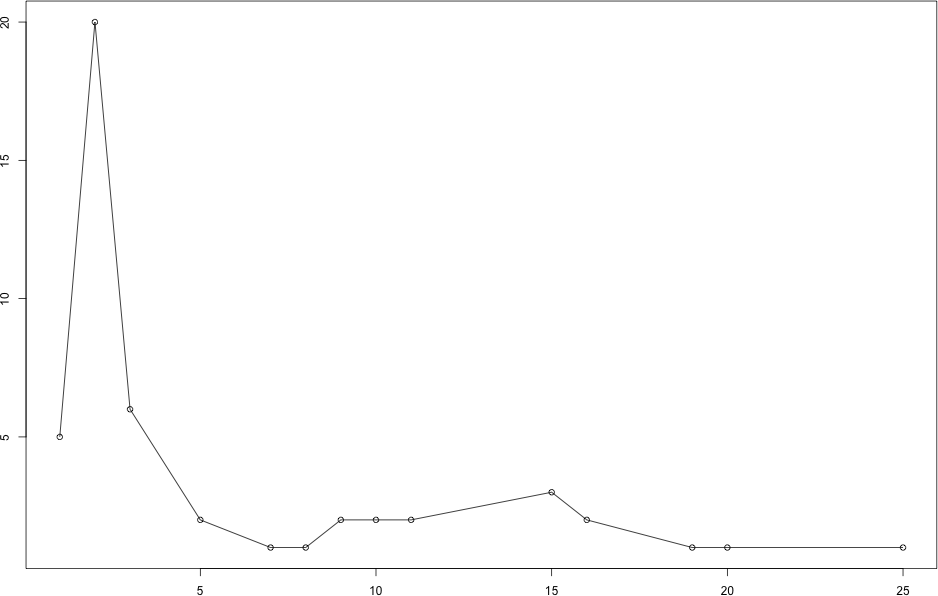}%
\label{fig14}}
\hfil
\subfloat[$\frac{LW_{ratio}}{A_{cumulative}}$ ]{
\includegraphics[width=2.5in]{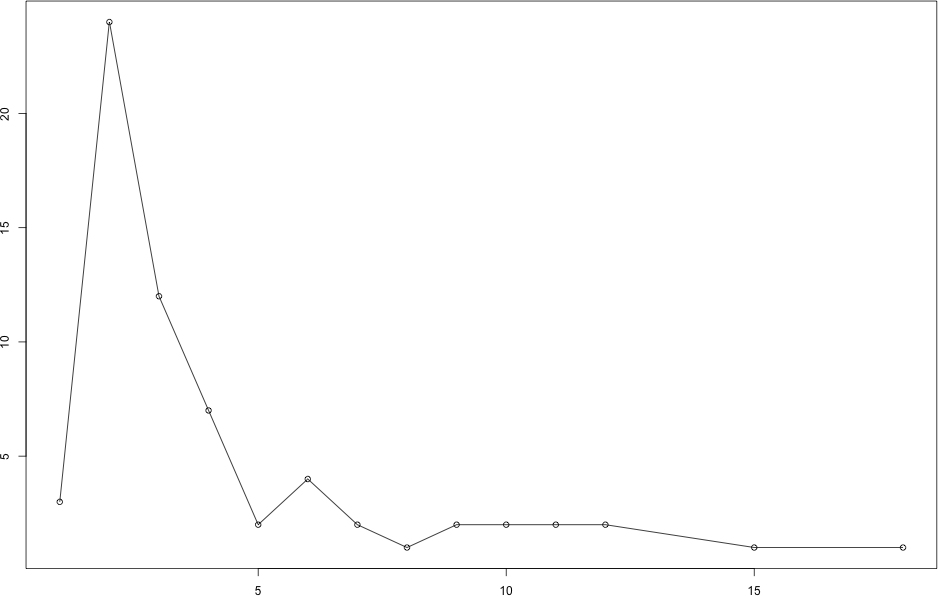}%
\label{fig15}}}
\caption{The histogram of break-point counts (the number of hierarchical elements with a particular count).}
\label{fig16}
\end{figure*}
\section{Candidate Ranking Procedure}
Supposing the hierarchical element $h_i$ is a (terminal) candidate cluster,
we rank candidate clusters by the following score, based on the coordinates of the representative pixels $P(h_i)$:
\[
S(h_i) = 
\frac{  \frac{1}{ |P(h_i)| } \underset{x \in P(h_i)}{\sum} |x_i - m(P(h_i)) |    }
{   \underset{x \in P(h_i) }{ \max }    |x_i - m(P(h_i)) |  }
\]
where we take the measure of centrality $m$ to be:
\[
	m(X) = \frac{1}{|X|} \underset{(x,y) \in X}{\sum} (x,y).
\]
The numerator is the 
Mean Absolute Deviation (MAD); the denominator is the Maximum Absolute Deviation. 
Then $S(h_i)$ represents a dispersion measure that is normalized by the object's scale (according to the Maximum Absolute Deviation).  Fig. 13 and Fig. 14 show the results ordered in increasing values of $S(h_i)$, for the termination according to  $A_{merge}$ and $\frac{LW_{ratio}}{A_{cumulative}}$ respectively.  Table I and II record the values of the dispersion $S(h_i)$ in both cases, for each terminal node $h_i$.
\begin{table}[!t]
\renewcommand{\arraystretch}{1.3}
\caption{Scores for $A_{merge}$ based termination candidates}
\label{table1}
\centering
\begin{tabular}{|c||c|c|c|}
\hline
$i(h_i)$ & MAD & Max.A.D.& $S(h_i)$ \\ \hline
161 & 1887 & 15.42 &122.3\\ \hline
150 & 4766 & 28.00 &170.1\\ \hline
226 & 4396 & 20.60 &213.3\\ \hline
82 & 17374 & 48.11 &361.0\\ \hline
72 & 27809 & 46.21 &601.7\\ \hline
27 & 27022 & 42.35 &638.0\\ \hline
\end{tabular}
\end{table}
\begin{table}[!t]
\renewcommand{\arraystretch}{1.3}
\caption{Scores for $\frac{LW_{ratio}}{A_{cumulative}}$ based candidates}
\label{table2}
\centering
\begin{tabular}{|c||c|c|c|}
\hline
$i(h_i)$ & MAD & Max.A.D. & $S(h_i)$ \\ \hline
106 &1958.4 &16.09 &121.6\\ \hline
161 &1887.7 &15.42 &122.3\\ \hline
225 &2983.9 &20.37 &146.4\\ \hline
23 &5886.5 &28.27 &208.1\\ \hline
138 &6922.4 &31.00 &223.2\\ \hline
100 &9353.1 &31.28 &299.0\\ \hline
14 &30478 &51.43 &592.5\\ \hline
\end{tabular}
\end{table}
\begin{figure}[!t]
\centering
\includegraphics[width=1.85in]{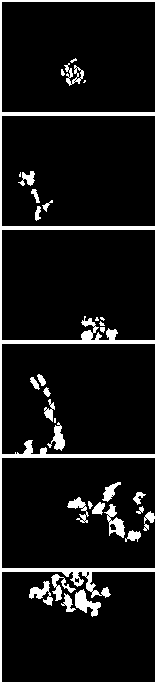}
\caption{ $A_{merge}$ based candidates ranked by $S(h_i)$}
\label{fig17}
\end{figure}
\begin{figure}[!t]
\centering
\includegraphics[width=1.85in]{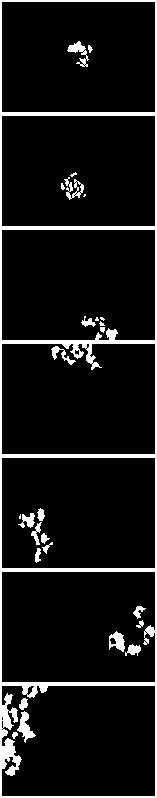}
\caption{ $\frac{LW_{ratio}}{A_{cumulative}}$ based candidates ranked by $S(h_i)$  }
\label{fig18}
\end{figure}
\section{Conclusion}
To comprehensively analyze the hierarchy of agglomeration, a framework for calculus along paths descending the agglomeration hierarchy was developed.  The framework accomplishes a locally adaptive termination procedure for hierarchical agglomerative clustering; this represents an alternative to conventional threshold termination rules based on fixing either the threshold distance, 
the number of clusters, or even the number of initial clusters, for example \cite{Lee2004} in the case of clustering for land-cover type discrimination using polarimetric SAR imagery.  Finally, a case study demonstrated the potential application of the new methodology 
to results of Gougeon's Individual Tree Crown (ITC) delineation\cite{gougeon}. 

Advancement of the framework in the context of multispectral and hyperspectral imagery 
 is a direction for further inquiry pertaining to ITC delineation, and to image analysis in general. 
The new termination used in conjunction with hierarchical clustering for multispectral imagery along with three dimensional visualization of the spatial and spectral extent of clusters in the spirit of \cite{richardson} could potentially lead to valuable 
interactive and visually intuitive insights into the spatial and spectral separability of land-cover types.

Beyond resource mapping, other potential applications include the analysis of atmospheric, astronomic, oceanographic, seismic, and medical imaging data.  Finally, we suggest that the cumulative extreme points associated with a hierarchy offer the possibility of
geometrically meaningful simplifications of hierarchies, for purposes of visualization or data compression.

\ifCLASSOPTIONcaptionsoff
  \newpage
\fi
\bibliographystyle{IEEEtran}
\bibliography{bib/clust}
\end{document}